\def\year{2020}\relax
\documentclass[letterpaper]{article} 
\usepackage{aaai20}  
\usepackage{times}  
\usepackage{helvet} 
\usepackage{courier}  
\usepackage[hyphens]{url}  
\usepackage{graphicx} 
\usepackage{graphicx}  
\usepackage{placeins}
\usepackage{hyperref}	
\usepackage{cleveref}

\title{Pre-training A Neural Language Model Improves The Sample Efficiency of an Emergency Room Classification Model}
\author{Binbin Xu\textsuperscript{\rm 1}, Cedric Gil-Jardine\textsuperscript{\rm 1,2}, Frantz Thiessard\textsuperscript{\rm 1}, \\
 \Large \textbf{Eric Tellier\textsuperscript{\rm 1,2}, Marta Avalos-Fernandez\textsuperscript{\rm 1,3}, Emmanuel Lagarde\textsuperscript{\rm 1}} \\
 \textsuperscript{\rm 1}University of Bordeaux, Bordeaux Population Health Research Center, UMR U1219, INSERM, F-33000, Bordeaux, France \\
 \textsuperscript{\rm 2}University Hospital of Bordeaux, Pole of Emergency Medicine, F-33000, Bordeaux, France \\
 \textsuperscript{\rm 3}SISTM team Inria BSO, F-33405, Talence, France \\ 
 {binbin.xu@mines-ales.fr}, \{first name.last name\}@u-bordeaux.fr}

\begin{document}

\maketitle

\begin{abstract}

To build a French national electronic injury surveillance system based on emergency room visits, we aim to develop a coding system to classify their causes from clinical notes in free-text. Supervised learning techniques have shown good results in this area but require a large amount of expert annotated dataset which is time consuming and costly to obtain. 
We hypothesize that the Natural Language Processing Transformer model incorporating a generative self-supervised pre-training step can significantly reduce the required number of annotated samples for supervised fine-tuning. 

In this preliminary study, we test our hypothesis in the simplified problem of predicting whether a visit is the consequence of a traumatic event or not from free-text clinical notes. Using fully re-trained GPT-2 models (without OpenAI pre-trained weights), we assess the gain of applying a self-supervised pre-training phase with unlabeled notes prior to the supervised learning task.    
Results show that the number of data required to achieve a ginve level of performance (AUC$>$0.95) was reduced by a factor of 10 when applying pre-training. Namely, for 16 times more data, the fully-supervised model achieved an improvement $<$1\% in AUC. To conclude, it is possible to adapt a multi-purpose neural language model such as the GPT-2 to create a powerful tool for classification of free-text notes with only a small number of labeled samples. 

\end{abstract}

\section{Introduction}
\subsection{The French emergency room surveillance system}
Syndromic surveillance is an approach in which automatic data recording procedures allow the provision of data for near-real-time tracking unexpected health events, monitoring expected trends and conducting health impact assessment of  infectious or environmental hazards.  
The French emergency room (ER) surveillance system (Oscour network) was implemented by the French National Public Health Agency in the early 2000s. 
Oscour ensures automatic and near real-time transmission of individual-level data and covers more than 90$\%$ of all emergency hospital visits in the country. The data quality concerning the demographic (date of birth, gender, location) and medical diagnoses (coded in International Classification of Diseases diagnostic codes, version 10, ICD-10) have permanently been increasing \cite{Fouillet2015French}. This surveillance system has therefore shown its utility in detecting and following-up public health events such  as seasonal outbreaks. 
However, the current lack of standardized data concerning the context and cause of the event makes epidemiological risk factor analysis unfeasible. In particular, injury epidemiology studies require the mechanism of traumatic injury to be documented: intentional/unintentional, self-inflicted/other-inflicted, resulting from a transport/workplace/home and leisure time accident, etc. 
Fortunately, the cause for the visit and injury mechanisms are fully described with free-text narratives stored in digital clinical records.
More than 21 million unlabeled ER clinical notes are produced every year in France open to potential exploitation.

\subsection{Pre-trained NLMs: an overview}
Over the past 10 years, neural language models (NLMs) have progressively taken the largest share in the field of natural language processing with techniques based on long short-term memory and gated recurrent networks \cite{Huang2019empirical} or convolutional networks \cite{Li2019Automated}. NLMs have then become indispensable in this field with applications like machine translation, document classification, text summarization and speech recognition.

The benefit of unsupervised pre-training have been quickly identified \cite{Erhan2010Why}, but in the domain of NLMs, new levels of performance have only been recently achieved with the use of models based on the concept of \emph{attention} that consists in learning dependencies between words in a sentence without regard to their distances. This mechanism has been implemented in a sequence to sequence neural network model, the Transformer architecture, proposed in 2017 \cite{Vaswani2017Attention}. This model can be trained with an unsupervised generative step that learns from a large set of text to predict the new token in a sentence \cite{Rothe2019Leveraging}. One of the latest examples is the OpenAI's Generative Pre-Trained Transformer-2 (GPT-2), published in February 2019. GPT-2 is a large transformer-based language model with 1.5 billion parameters, trained on a dataset of 8 million web pages to predict the next word after a given prompt sentence \cite{Radford2019Language}. This work quickly drew attention from the community as it demonstrated the model's ability to generate artificial texts which are difficult to be distinguished from  humans written texts. Moreover, the meaning of these artificial sentences was surprisingly consistent with the original context text (prompt), suggesting potentially numerous applications. 
Indeed, beyond the capability to generate coherent texts, fine-tuning the GPT-2 model has the potential to perform other tasks such as question answering and document classification. Following the same idea as the Bidirectional Encoder Representations from Transformers (BERT) model \cite{Devlin2018BERT}, transferring many self-attention blocks from a pre-trained model proved sufficient to transfer contextual representations in the dataset.

The training of the model is then performed in two distinct phases \cite{Radford2018Improving}: the first generative pre-training unsupervised (or more accurately \emph{self-supervised}) phase, consists in exploitation of a text corpus. This leads to the ability of automatic text generation. The relevance of these synthetic sentences suggests  that the networks learned contextual semantic representations. The second supervised fine-tuning phase consists in resuming learning from annotated text corpus with the objective of creating a system able to perform specific tasks.

\subsection{Objective}
 The so-called  TARPON  project proposes to build a French national surveillance system based on the exhaustive collection of ER visits reports in France for automated trauma monitoring. Its main feature is the application of automatic language analysis to extract injury mechanism and cause from the digital medical record of each ER visit. 
The overall objective is to develop a tool that would derive standardized trauma/injury information and their causes from these ER notes. To that purpose, substantial amounts of experts-annotated data would be necessary to train a conventional text classification model with acceptable accuracy.

We hypothesize that the GPT-2 incorporating a generative self-supervised pre-training step can significantly reduce the required number of expert annotated samples for supervised fine-tuning. 
This is of paramount significance for all projects wishing to use NLMs models for free-text classification tasks because the manual annotation phase is by far the most expensive one. The objective of the present study is to measure the gain in terms of manual annotation load obtained by adopting this pre-training step.

\section{Methods}

\subsection{Study design and data sources}

To test our hypothesis and measure the gain, we exploited the current digital medical record data of our ER department. 
We leveraged the fact that the traumatic/non-traumatic cause of the ER visit 
could be easily derived from available diagnostic codes (ICD-10) already assigned by clinicians at patient's hospitalization. 
We also retrieved clinical notes from the digital medical record system of the adult ER of our University hospital, from 2011 to 2018.  The ICD-10 \cite{Organization2015International} is the most used standard way to indicate diagnoses and medical procedures, and is the mandatory terminology used in France for all stays in any private or public hospitals. 
This data set then contains $288\,404$ medical records of which $209\,341$ contain both diagnosis code (from which we derived the traumatic/non-traumatic cause) and the free-text clinical note.
A total of $56\,410$ visits with ICD-10 codes beginning with letters S, T1 to T35 and V were coded as trauma and $115\,520$ visits with ICD-10 codes beginning with letters A, C, D, E, G, H, I, J, L, N were coded as non-trauma. A total of $37\, 411$ visits (with codes beginning with letters F, M, O, P, Q, T36 to T98, X40 to X57, Y10 to Y98, U, Z) were excluded since they correspond to pathologies for which the traumatic nature is either uncertain or discussed from a semantic point of view. The total number of available clinical notes was therefore $171\,930$.

We then trained the GPT-2 model to predict from free-text clinical notes only whether the visit was due to a traumatic VS non-traumatic cause.
We designed the following two scenarios to predict if an ER visit was due to trauma or not from free-text clinical notes and to measure the gain obtained with a self-supervised pre-training phase. 
Scenario A consisted in retraining the GPT-2 from scratch on the labeled dataset in a single fully-supervised phase. In Scenario B, we further split the training dataset in two parts: a large unlabeled dataset for the self-supervised pre-training phase and a smaller labeled dataset for the supervised training. The main question was therefore how many clinical notes were required in this training part of Scenario B to achieve the same acceptable performance as in Scenario A. This should give us a measure of how much annotation load can be gained. 

\subsection{Sampling strategy}

\begin{figure}[!htb]
\centering
\includegraphics[width=.9\columnwidth]{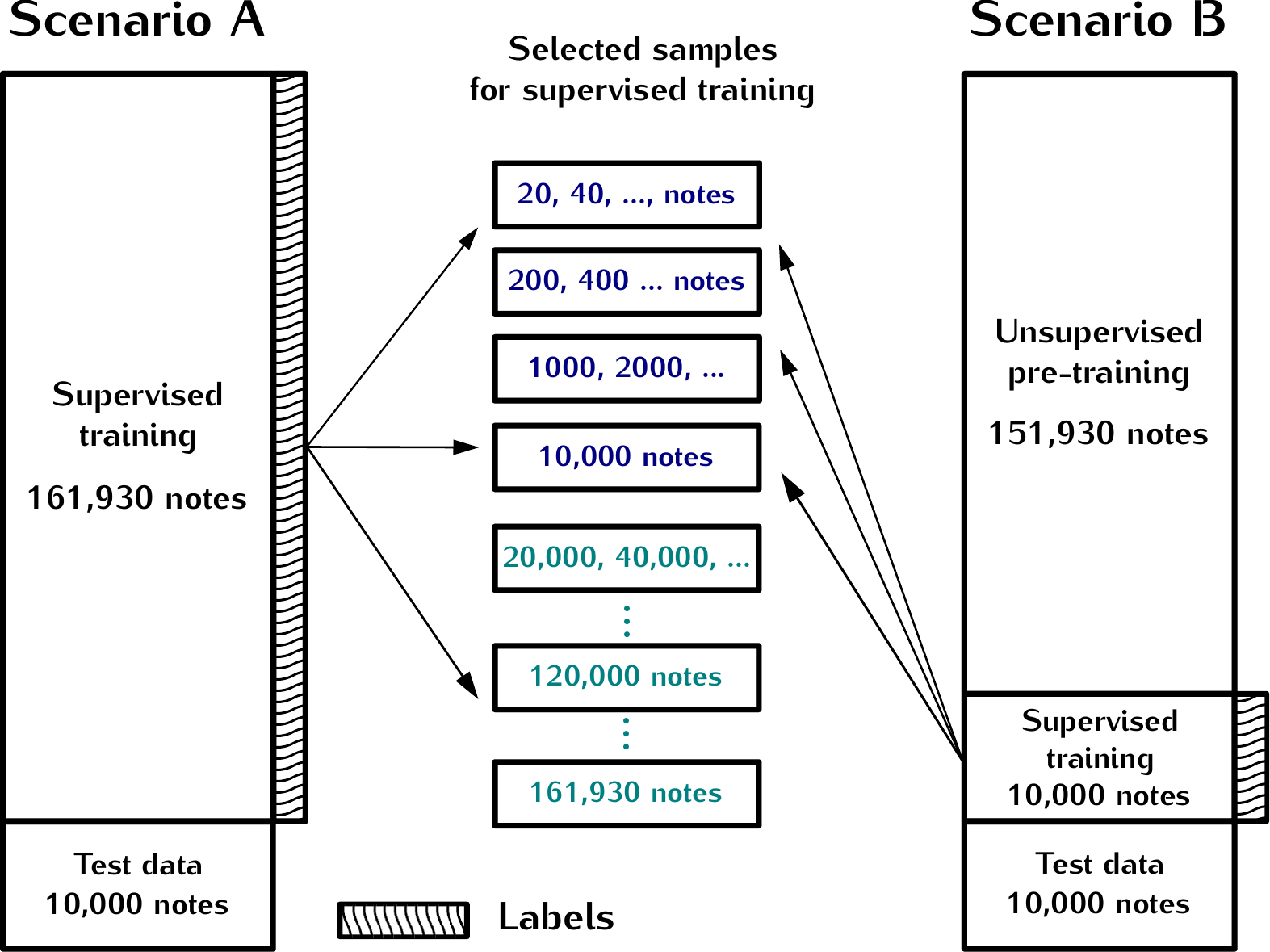}
\caption{Strategy of sampling: $26$ cases evaluated in {\scshape{Scenario A}} and $19$ cases in {\scshape{Scenario B}}.}
\label{figure_03}
\end{figure}

The sampling strategy is illustrated on \Cref{figure_03}. For test purposes, $10\,000$ clinical notes were randomly selected and then frozen for both scenarios. This test set was used to estimate the number of notes needed to achieve maximum prediction performance. 
The clinical notes from the remaining $161\,930$ notes were used with labels in Scenario A. For Scenario B, we further split the $161\,930$ notes into a set of $151\,930$ unlabeled notes for unsupervised pre-training and a second set with $10\,000$ labeled notes for the supervised fine-tuning step. 

Models were independently trained with different numbers of notes. We built a grid of values of numbers of notes: 
\begin{itemize}
    \item  19 cases regularly spaced on a log-scale, from $20$ to $10\,000$, shared by the two scenarios; 
    \item plus 6 cases regularly spaced on a log-scale, from $20\,000$ to $120\,000$, plus  $161\,930$, the total number of notes, only  pertinent for Scenario A (indeed, in Scenario B,  only   $10\,000$ notes are used for supervised learning on labeled notes thus, only the first 19 cases could be done).
\end{itemize}
 The number of cases to be evaluated was chosen in agreement with our computing capabilities.

\subsection{Adapting the  GPT-2  model to our setting}

Like other NLMs based on convolutional  and recurrent neural networks, the GPT-2 proposed by Radford and colleagues is a sequence to sequence transduction model \cite{Cho2014Learning}. The main feature of the Transformer architecture is to use \emph{attention} weight on text inputs  \cite{Vaswani2017Attention}. During the training process, the network learns a context vector which gives global level information on inputs telling where {attention} should be focused. The novel approach consists in replacing recurrence with {attention} to handle the dependencies in input and output.

The GPT-2 is built to predict the next token from the input of a text sequence. By looping this process, it works then as a text generator. The text can be generated de novo or by feeding any arbitrary text prompts. The model was originally trained on millions of webpages without any explicit supervision. Four models of GPT-2 with respectively 117, 345, 762 and 1542 million parameters were trained. Only the first two models are trainable on standard workstations with an appropriate Nvidia GPU.
Note that the GPT-2 models are trained with web text mostly written in English while our clinical notes data are all in French. Consequently in the present work, we did not use those pre-trained models and retrained the models from a random set of weights.

The 117M models were trained mainly with a single Nvidia\textsuperscript{\textregistered} GeForce GTX 1080 Ti with 11GB of VRAM (4 parallel sessions can be run on our workstation with 4 GTX 1080 Ti). The 345M models were trained on another workstation with a single Nvidia\textsuperscript{\textregistered} TITAN RTX with 24GB of VRAM.

\subsection{Operating principle}

In Scenario B, the pre-training step is referred as unsupervised learning because it is derived from simply reading the unlabeled clinical notes. It actually uses a sliding learning window on the text. The first part of this window corresponds to the input and the last token is then the token to be predicted. This first step leads to models that can generate texts resembling clinical notes in French, including the use of medical jargon and specialized abbreviations.

For the supervised learning phases (Scenario A and second training process in Scenario B), we added the string ``TARPON'' to the end of each clinical note and the next position of string, ``1'' (if the clinical note corresponds to a traumatic event) or ``0'' (if non-traumatic event), is used for training label and to be predicted. As described above, this code was derived from the diagnosis classification manually coded by clinicians.

\begin{figure}[!htb]
\centering
\includegraphics[width=.9\columnwidth]{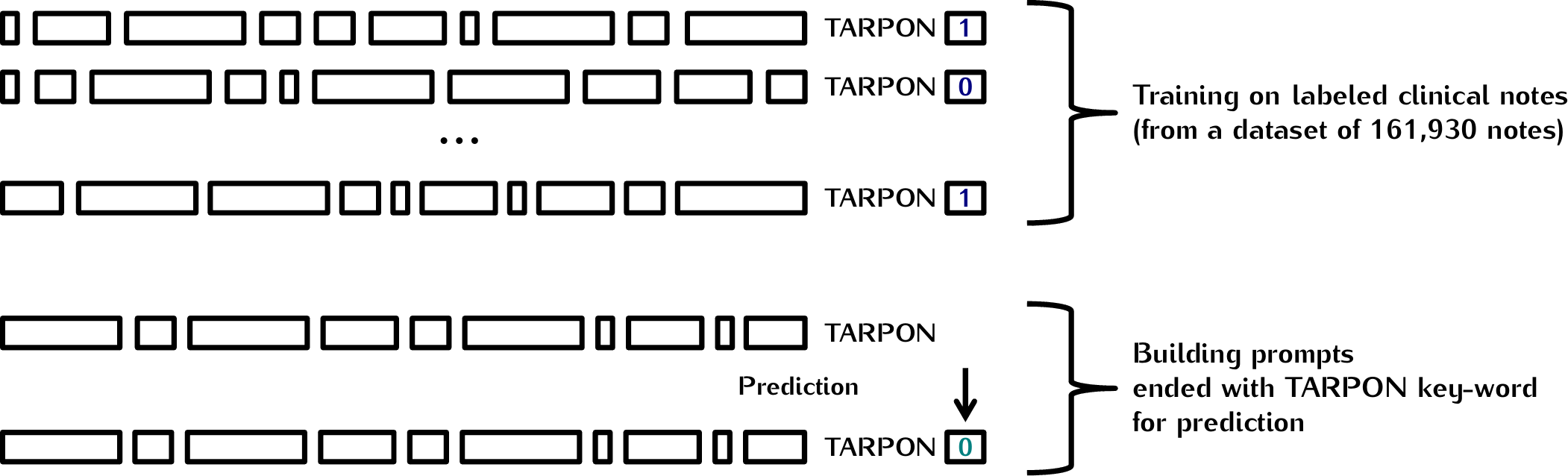}
\caption{{\scshape{Scenario A}}: supervised training}
\label{figure_01}
\end{figure}

\begin{figure}[!htb]
\centering
\includegraphics[width=.9\columnwidth]{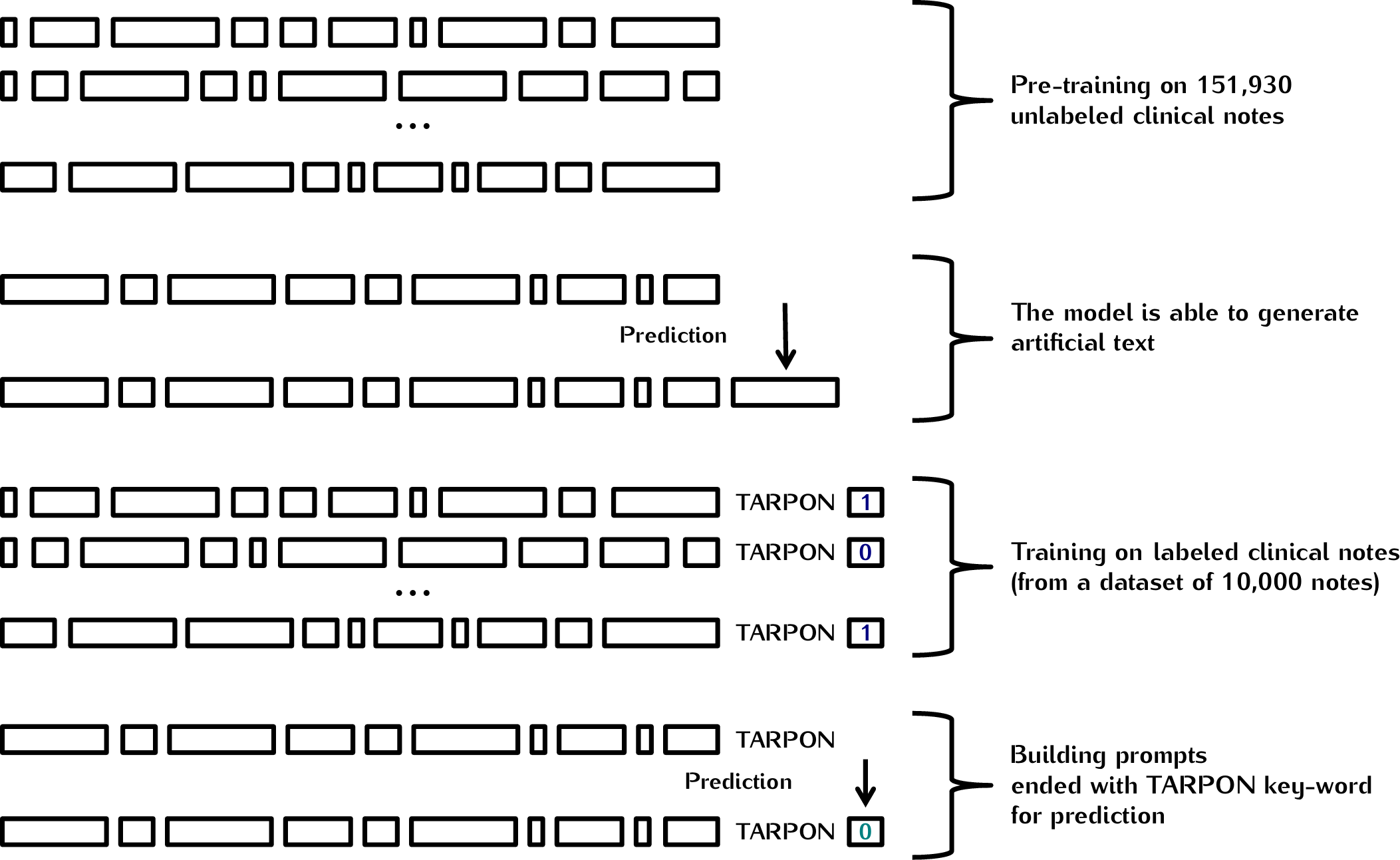}
\caption{{\scshape{Scenario B}}: self-supervised training + supervised training}
\label{figure_02}
\end{figure}

For both scenarios (\Cref{figure_01,figure_02}), the test phase consists in feeding the models with prompts by adding the task identifier at the end of each test clinical note and ask the model to predict the next token right after the task identifier. Ideally, this newly generated token should be one of the classification codes (tokens). On the first iterations, due to the random initialization and insufficient learning, the predicted token could be any tokens from the vocabulary other than expected classification tokens. But, this turns quickly to be mainly the classification tokens. Our operating principle can therefore be compared to a Question Answering task.

The prediction performance of the model was measured by {F1 score} and area under the ROC curve statistics ({AUC}) \cite{Powers2011Evaluation}. Evaluations on the same $10\,000$ clinical notes were performed for both Scenario A and Scenario B.

\subsection{Confidentiality and data protection}

No nominative data were necessary for this work. The dataset was not checked and not specifically deidentified. Data processing and computing were conducted within the facilities of the Emergency department of the Bordeaux University Hospital which have received regulatory clearance to host and exploit databases with personal and medical data.

\section{Results}

\begin{figure}[!ht]
\centering
\includegraphics[width=.9\columnwidth]{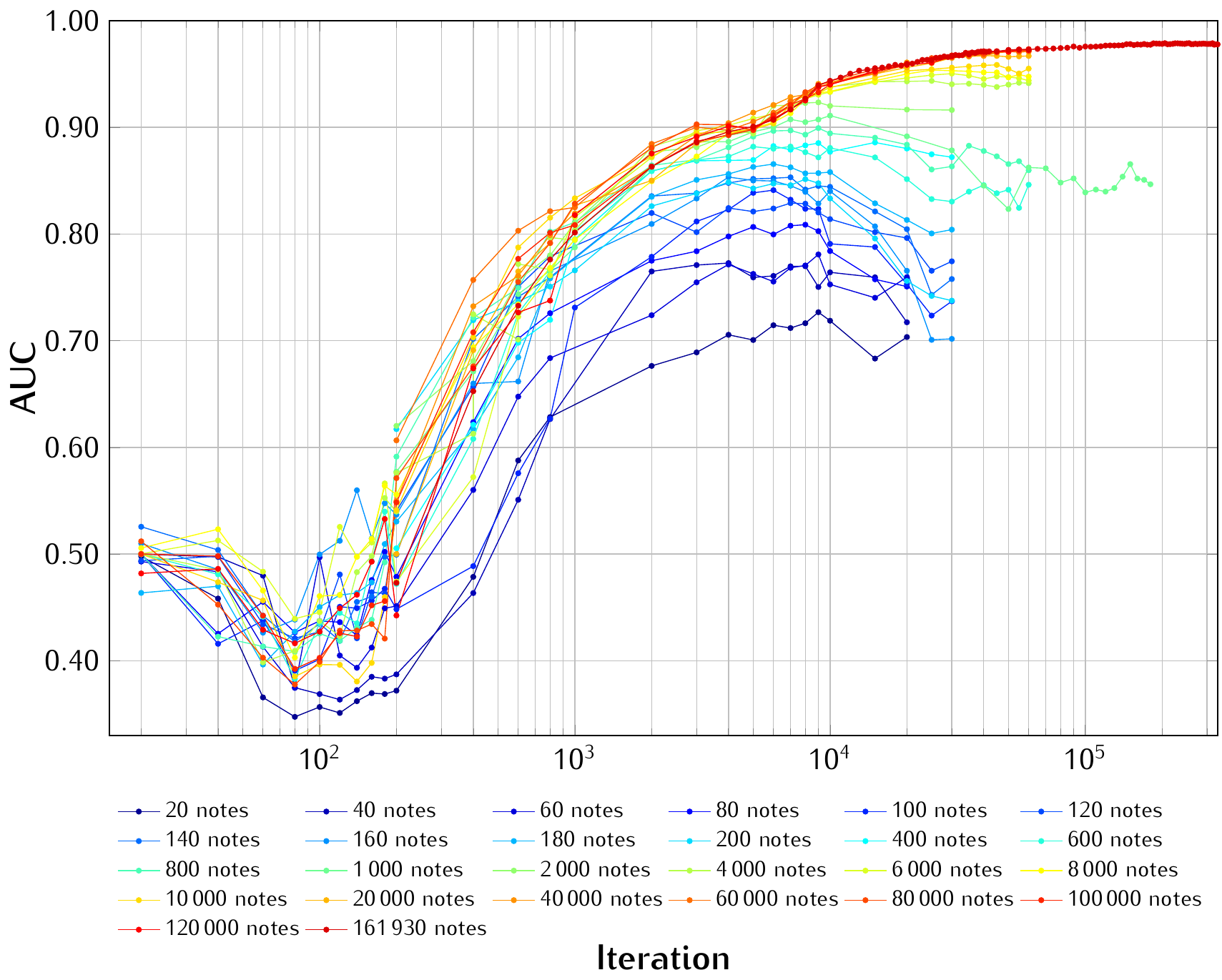}
\caption{Scenario A: {AUC} by number of iterations. 26 cases.}
\label{figure_04}
\end{figure}

\begin{figure}[!ht]
\centering
\includegraphics[width=.9\columnwidth]{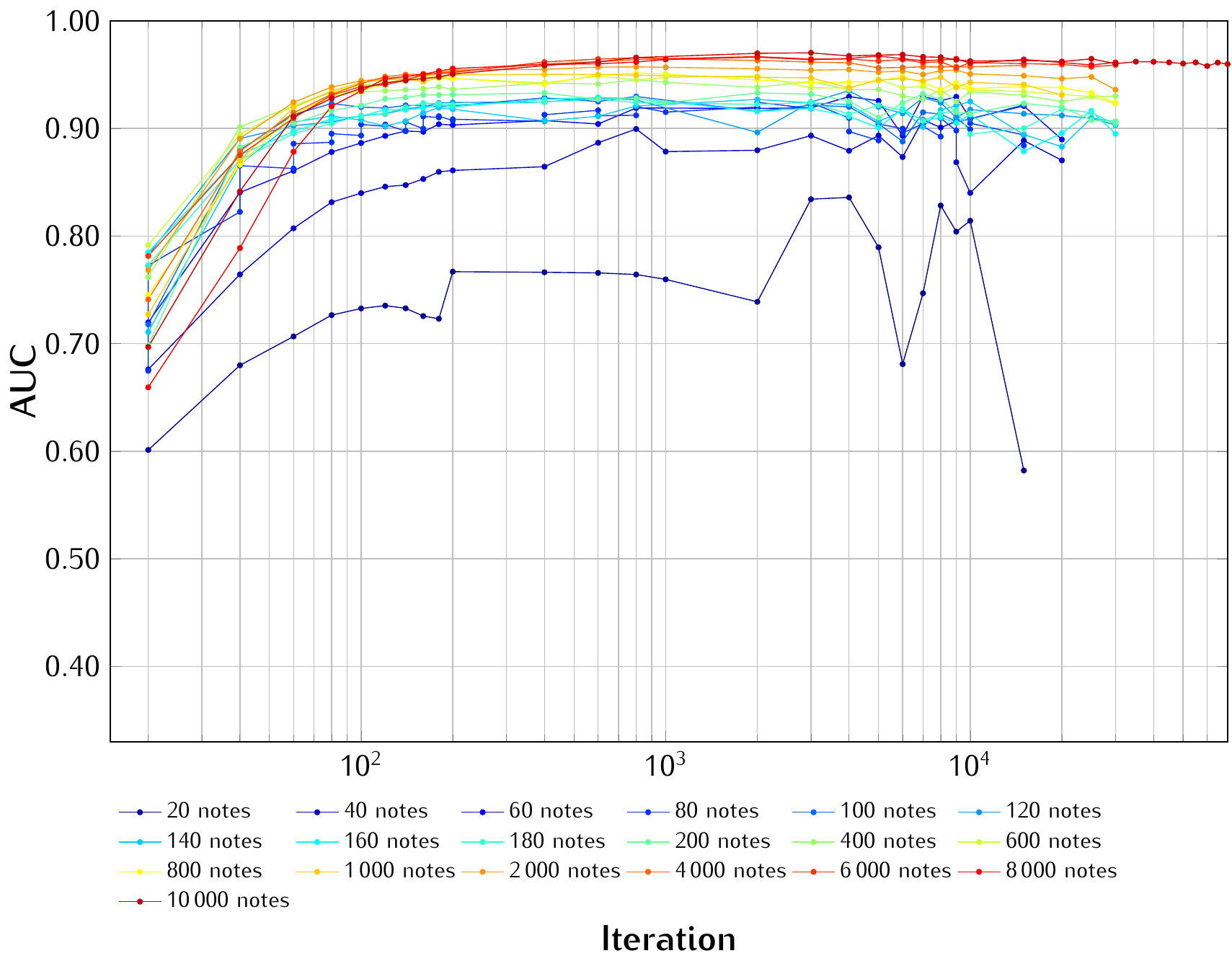}
\caption{Scenario B: {AUC} by number of iterations. 19 cases.}
\label{figure_05}
\end{figure}

\begin{figure}[!ht]
\centering
\includegraphics[width=.9\columnwidth]{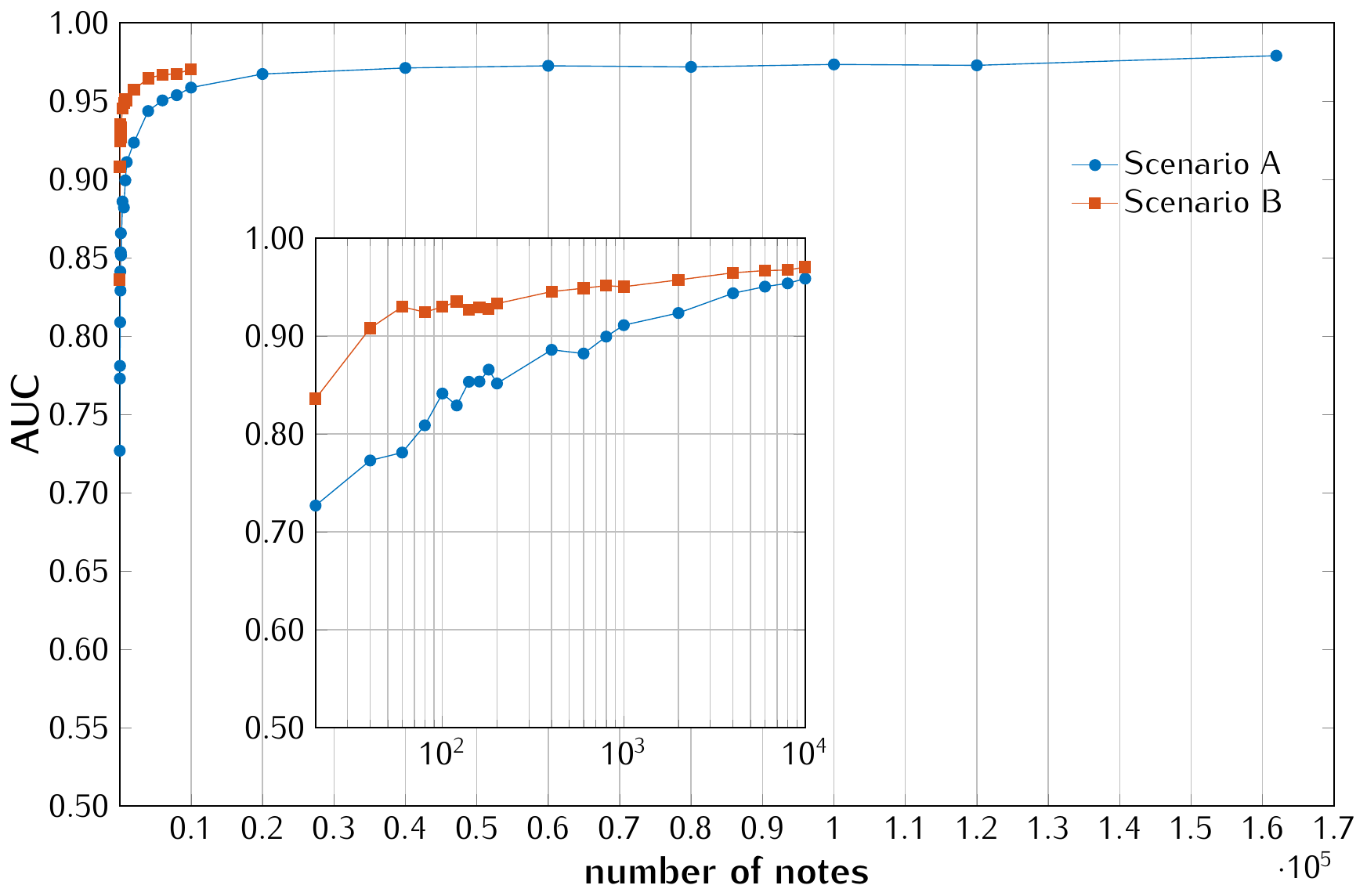}
\caption{Comparison of {AUC} by cases (number of notes) for both scenarios}
\label{figure_06}
\end{figure}

\begin{figure}[!ht]
\centering
\includegraphics[width=.9\columnwidth]{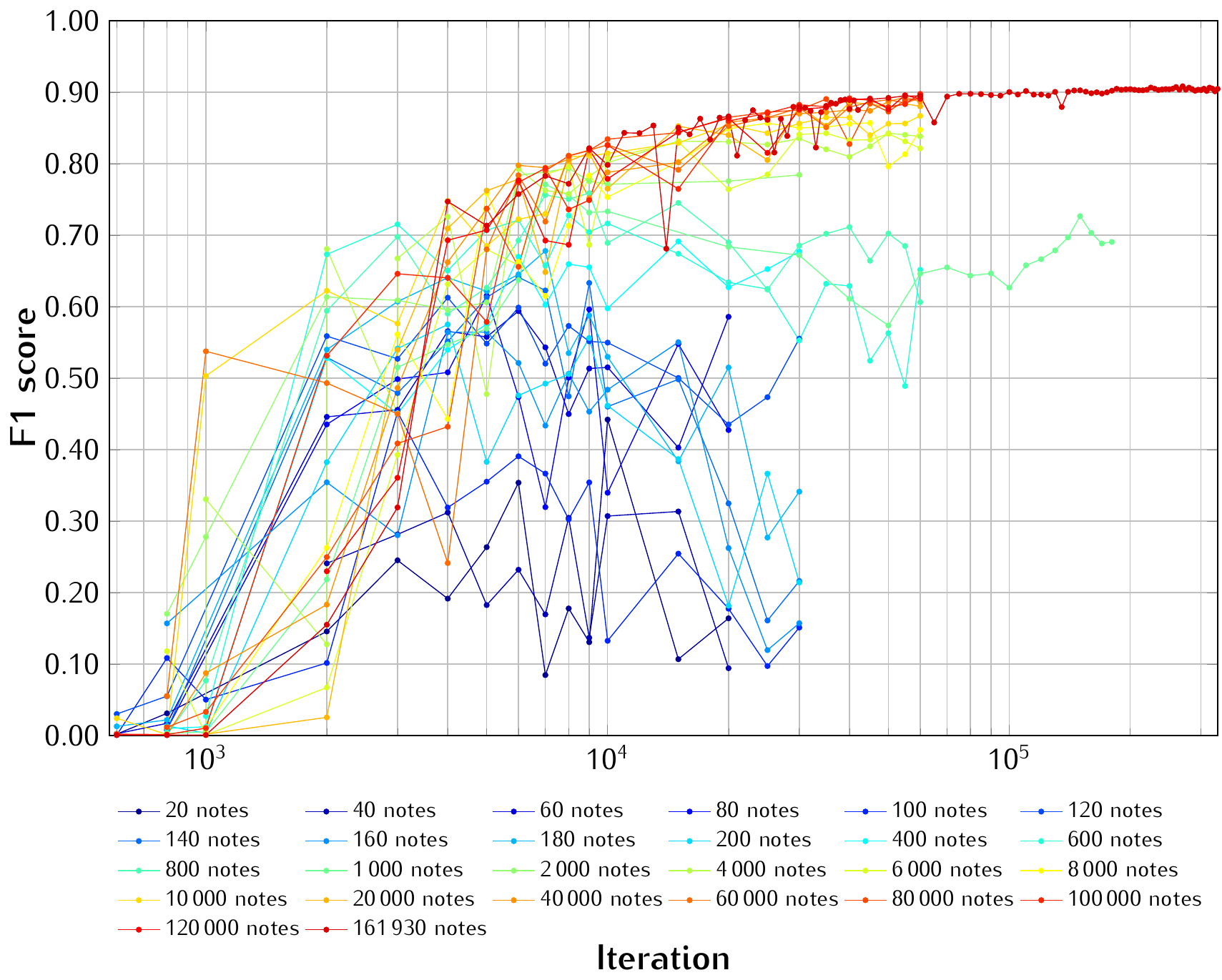}
\caption{Scenario A: {F1 score} by number of iterations. 26 cases.}
\label{figure_07}
\end{figure}

\begin{figure}[!ht]
\centering
\includegraphics[width=.9\columnwidth]{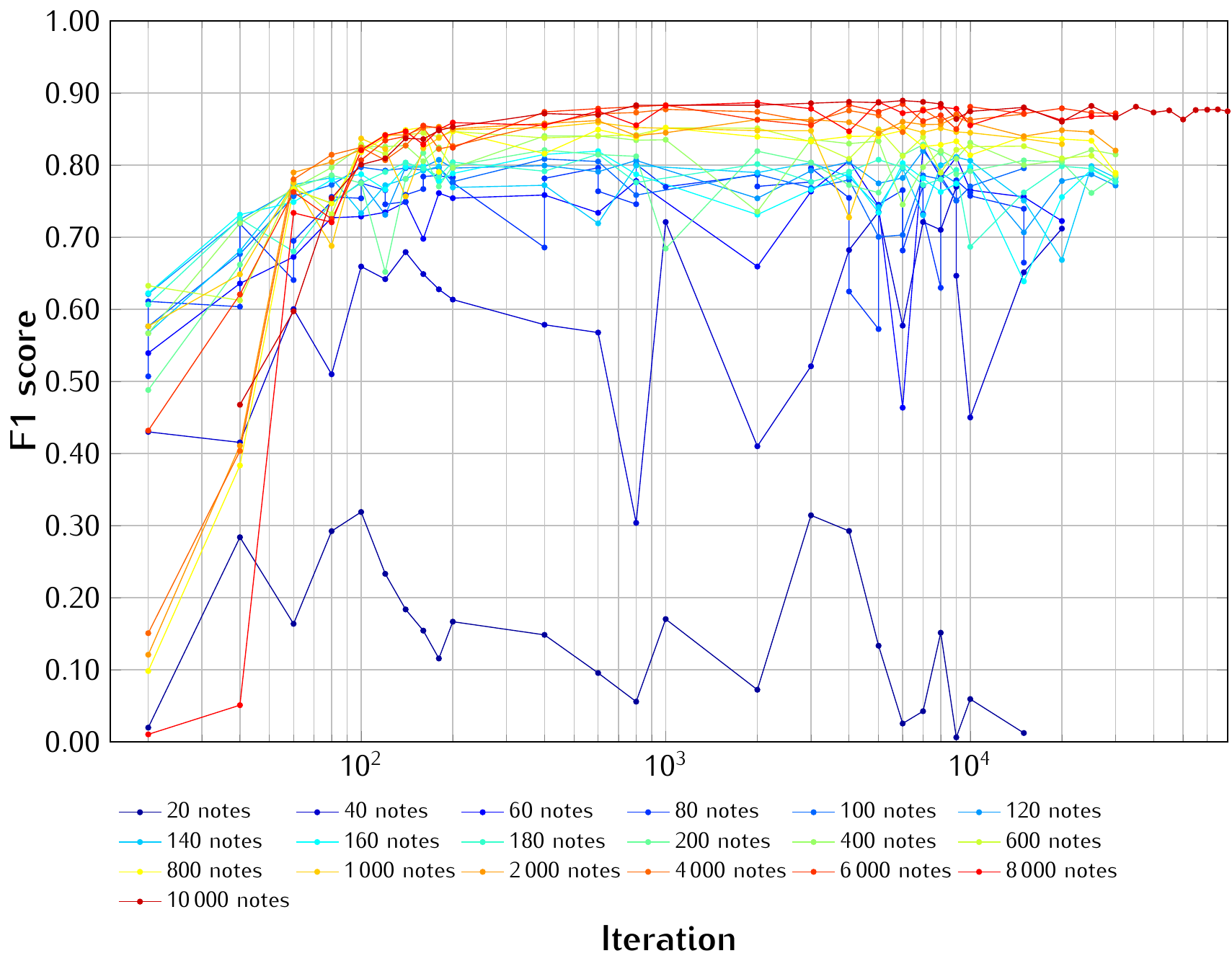}
\caption{Scenario B: {F1 score} by number of iterations. 19 cases.}
\label{figure_08}
\end{figure}

\begin{figure}[!ht]
\centering
\includegraphics[width=.9\columnwidth]{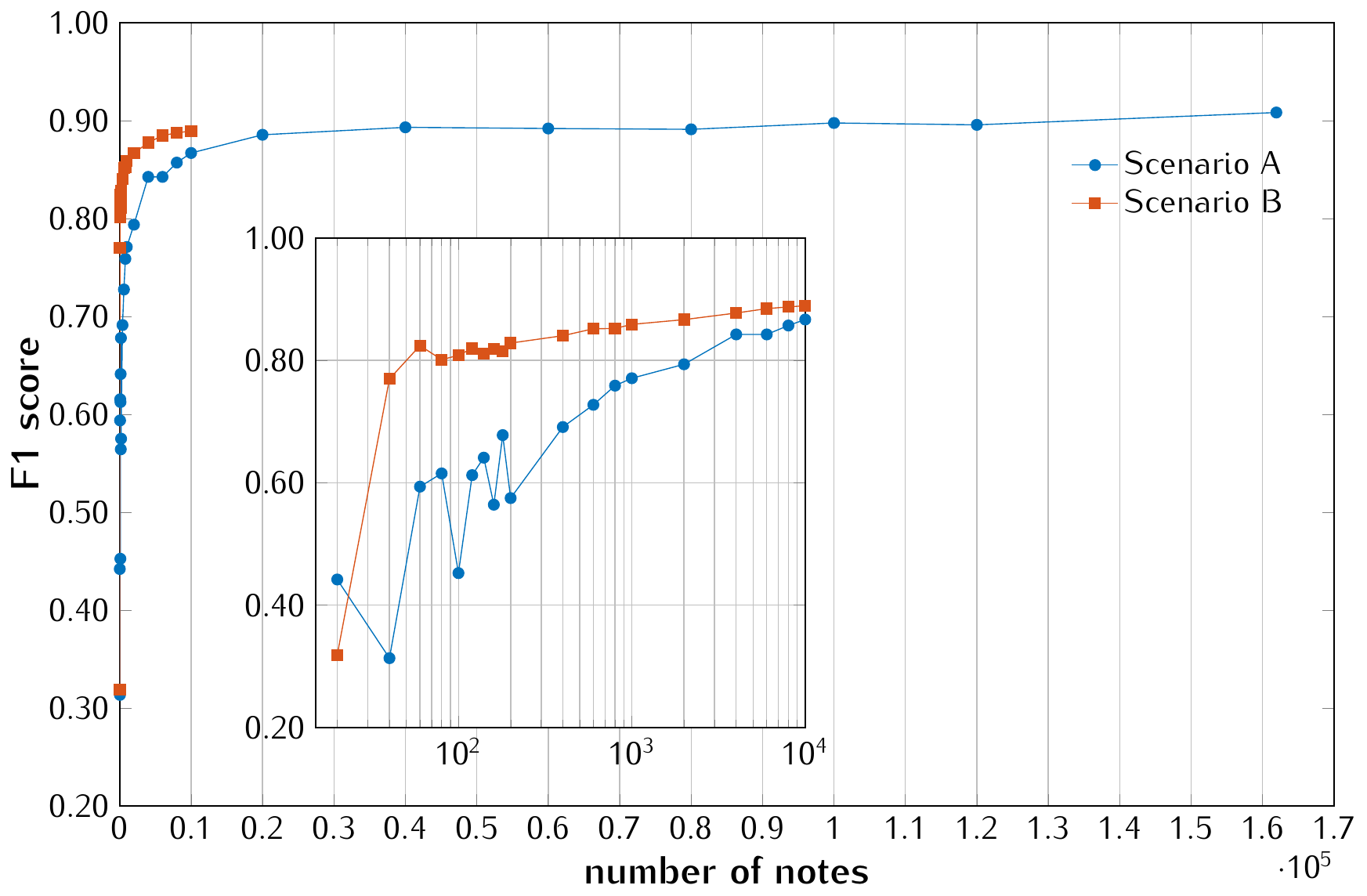}
\caption{Comparison of {F1 score} by cases (number of notes) for both scenarios}
\label{figure_09}
\end{figure}

For both scenarios, we compared {AUC} (\Cref{figure_04,figure_05,figure_06}) and {F1 score} (\Cref{figure_07,figure_08,figure_09}) by iterations with a batch size of 1. 
The number of iterations needed to achieve a maximal {AUC}/{F1 score} value varied depending on the number of notes (\Cref{figure_04,figure_05} for {AUC} and \Cref{figure_07,figure_08} for {F1 score}). For each set of clinical notes, the maximum {AUC}/{F1 score} value was retained to build \Cref{figure_06} for {AUC} and \Cref{figure_09} for {F1 score}, thus representing how model performance varied with respect to the number of labeled notes.

In Scenario A, {AUC} and {F1 score} reach the values of $0.979$ and $0.908$ respectively after the processing of all the $161\,930$ labeled notes. The use of generative pre-training (Scenario B) achieved an {AUC} of $0.949$ and an {F1 score} of $0.852$ after the processing of only $600$ labeled clinical notes. To achieve the same performance, $6\,000$ labeled clinical notes had to be processed in Scenario A (\Cref{figure_06,figure_09}).
At the end of Scenario B, with a training of all $10\,000$ notes, {AUC} and {F1 score} are respectively $0.970$ and $0.889$, corresponding the cases of more than $100\,000$ notes in Scenario A. 
For 16 times more data, the gain from Scenario A compared to Scenario B shows  an improvement of only $0.89\%$ in {AUC} and $2.12\%$ in {F1 score}.

The same is observed for the {F1 score}. In Scenario A (\Cref{figure_07}), the {F1 score} cannot be measured for the first 500 iterations since recall and precision are both null. While for Scenario B, {F1 score} can be measured after only 20 iterations (up to $0.633$) and reached $0.878$ with 600 iterations. For comparison, in Scenario A, the {F1 score} was only around $0.03$ after 600 iterations.

As regard to training time, one iteration (batch size was 1) took about $0.25$ second; the required number of iterations depended on the data length and varied from $15\,000$ to $330\,000$ which resulted in training time ranging from 1 to 23 hours. The prediction task on the $10\,000$-notes dataset lasted around 4 minutes for each iterations. As a result, the time for each case run took from 4 hours up to 100 hours.

Comparing 117M and 345M GPT-2 models showed no significant improvement using a more complex model (\Cref{{figure_10}}). However, the 345M model takes around $1.5$ second for each iteration ($6\times$ longer). The classification task of $10\,000$ notes with 345M model required about $480$ seconds which is $2\times$ longer than with 117M model ($245$ seconds). 
Considering the time cost and performance, all the above-mentioned results (\Cref{figure_04,figure_05,figure_06,figure_07,figure_08,figure_09}) are trained with the GPT-2 117M model. 

\begin{figure}[!htb]
\centering
\includegraphics[width=\columnwidth]{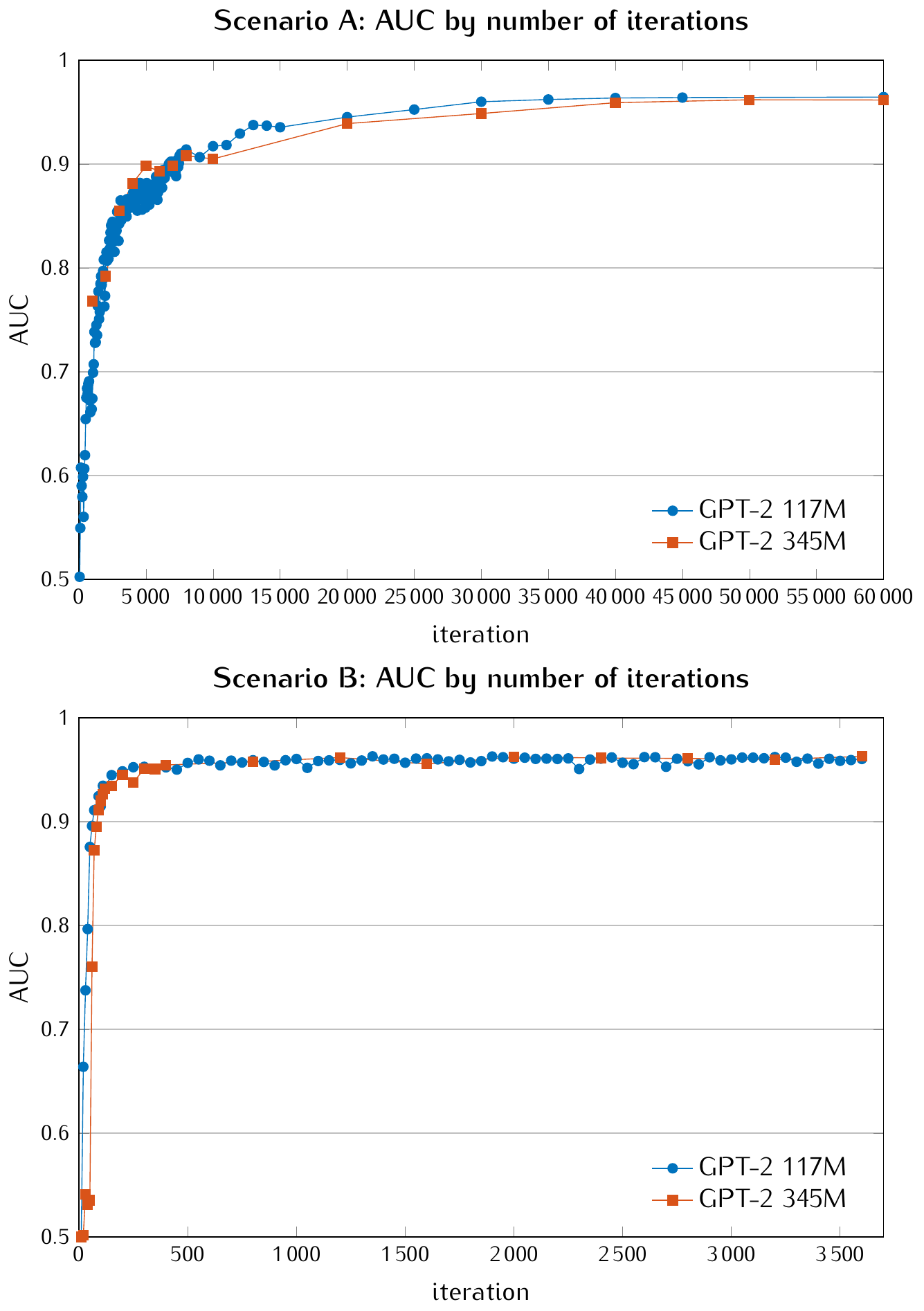}
\caption{Comparison of GPT-2 117M models and 345M model on the cases of $161\,930$ notes in Scenario A and $10\,0000$ notes in Scenario B.}
\label{figure_10}
\end{figure}

\section{Discussion}

As suggested by Radford and colleagues \cite{Radford2018Improving}, large gains could be obtained by generative pre-training with unlabeled text corpus, saving a large amount of annotation load. In our example of clinical notes classification task, the order of magnitude is a factor of 10. In their 2019 paper, Radford and colleagues reported an improvement of $8.9\%$ on commonsense reasoning (Stories Cloze Test), $5.7\%$ on question answering (RACE), and $1.5\%$ on textual entailment (MultiNLI) \cite{Radford2018Improving}.

Though the {AUC} converged to the same ending point in both scenarios, the learning patterns were quite different. In the fully-supervised scenario, the {AUC} started with a value $\sim 0.5$. Because of insufficient learning, almost all clinical notes were classified as non-trauma at this stage. The {AUC} dropped during the first iterations due to clinical notes wrongly classified as trauma, then increased as expected. 
The main reason is that, in this question answering design, the model has to perform two tasks at the same time: how to learn the semantic representation in clinical notes and how to perform the classification task. But for the pre-trained scenario, the clinical notes generation task is learned during the pre-training phase, leading to an increasing monotonous {AUC} curve  in step 2, corresponding to the learning of the classification task.

Our results are in line with recent work using self-supervised pre-training methods, such as ELMo \cite{Peters2018Deep} and BERT \cite{Devlin2018BERT}, and have established a qualitatively new level of performance in most widely used Natural Language Understanding benchmarks. Howard and Ruder \cite{Howard2018Universal} in particular reported very similar results in a comparable text classification task, with a model trained with only 100 labeled samples that matches the performance of training from scratch on $20\,000$ samples. While the extensive use of pre-trained word embeddings could be considered as of the same nature of generative pre-training, the gain provided by generative pre-training is a major step for those who seek to classify free-text document with minimal manual coding efforts for same acceptable accuracy.

We have benefited from the work of the researchers who published the GPT-2 model, which still seems to be one of the most efficient today. The NLM field progresses fast with extensive research efforts from the community. Other models have been and will be proposed, so the text classification strategies will need to be updated. Recent and promising work includes the work of Yang and colleagues and their XLNet model \cite{Yang2019XLNet} which currently ranks first at the Standford Question Answering Dataset (SQuAD2.0).

Probably because the GPT-2 model was only recently made public, few applications have been published today. However, this type of tool will with no doubt be extensively used in the near future for a wide range of tasks. In the area of document classification alone, they will likely provide faster and more relevant access to expected information. Certainly, these applications will go beyond simple classification tasks. Of note, it is unusual to generate the next token (in a Question Answering fashion) in an NLP model to perform classification tasks. A more classical approach would certainly be to add a soft-max layer after a hidden state of the model  to output prediction probabilities. While this will be done in future work, adding a layer however requires much more skill in Python/TensorFlow programming. That is why we decided to present a method that can be used by a much broader scientific community. 

While the 345M GPT-2 model did not generate better results than the 117M model in the current study, the use of larger models could bring further improvement. Unfortunately, the required computing power of larger models is far beyond our resources for this pilot study, we will have to be satisfied with the results presented here.

In this study, the trauma/non-trauma labeling procedure of the clinical notes was indirectly based on the ICD-10 codes. We tried to maximize the consistency of the ground-truth labeling by selecting a subset of ICD-10 codes for which the traumatic/non-traumatic characteristic is indisputable. 
This method has had the advantage of providing a large amount of labeled data but does not allow us to compare the model's performance with human annotation.

\section{Conclusion}

Our work shows that it is possible to easily adapt a multi-purpose NLM model such as the GPT-2 to create a powerful classification tool of free-text notes even in languages other than English. The self-supervised training phase appeared to be a very powerful tool to dramatically decrease the number of labeled samples required for supervised learning. 

Our results could be refined by extending the experiment to multi-label classification of ICD-10 codes. 
In the coming months, based on the results obtained, the exhaustive coding of all events leading to trauma with emergency room visits will be implemented. The multilingual aspect of the problem (clinical notes are in French and include clinical jargon, slang words, abbreviations, acronyms, and shortcuts) deserves further investigation. Deidentification questions should also be explored. 

These are the first steps towards a national trauma observatory within the TARPON project framework. More generally, this also opens broad perspectives for those interested in automatic free-text annotation. In the field of health, this will be particularly useful for diagnosis coding, clinical report classification and patient reports analysis and mining.

\fontsize{9.0pt}{10.0pt}
\selectfont
\bibliographystyle{aaai}  
\bibliography{bib}

\end{document}